\documentclass{article} 
\usepackage{nips12submit_e,times}
\usepackage{graphicx,url,amsmath}

\def\BE{\vspace{-0.0mm}\begin{equation}}
\def\EE{\vspace{-0.0mm}\end{equation}}

\def\BEA{\vspace{-0.0mm}\begin{eqnarray}}
\def\EEA{\vspace{-0.0mm}\end{eqnarray}}

\newcommand{\eqn}[1]{Eqn.~\ref{eqn:#1}}
\newcommand{\fig}[1]{Fig.~\ref{fig:#1}}
\newcommand{\tab}[1]{Table~\ref{tab:#1}}
\newcommand{\secc}[1]{Section~\ref{sec:#1}}
\def\etal{{\textit{et~al.~}}}

\title{Stochastic Pooling for Regularization of \\ Deep Convolutional Neural Networks}

\author{
Matthew D. Zeiler \\
Department of Computer Science \\
Courant Institute, New York University \\
\texttt{zeiler@cs.nyu.edu} \\
\And
Rob Fergus \\
Department of Computer Science \\
Courant Institute, New York University \\
\texttt{fergus@cs.nyu.edu} \\
}

\nipsfinalcopy 

\begin{document}

\maketitle

\begin{abstract}
  We introduce a simple and effective method for regularizing large
  convolutional neural networks. We replace the conventional
  deterministic pooling operations with a stochastic procedure,
  randomly picking the activation within each pooling region according
  to a multinomial distribution, given by the activities within the
  pooling region. The approach is hyper-parameter free and can be
  combined with other regularization approaches, such as dropout and
  data augmentation. We achieve state-of-the-art performance on four
  image datasets, relative to other approaches that do not utilize
  data augmentation.
\end{abstract}

\section{Introduction}

Neural network models are prone to over-fitting due to their high
capacity. A range of regularization techniques are used to prevent
this, such as weight decay, weight tying and the augmentation of the
training set with transformed copies \cite{NNtricks}. These allow the
training of larger capacity models than would otherwise be possible,
which yield superior test performance compared to smaller un-regularized models.

Dropout, recently proposed by Hinton \etal \cite{Hinton12}, is another
regularization approach that stochastically sets half the activations
within a layer to zero for each training sample during training. It
has been shown to deliver significant gains in performance across a
wide range of problems, although the reasons for its efficacy are not
yet fully understood.

A drawback to dropout is that it does not seem to have the same
benefits for convolutional layers, which are common in many networks
designed for vision tasks.  In this paper, we propose a novel type of
regularization for convolutional layers that enables the training of
larger models without over-fitting, and produces superior performance
on recognition tasks.

The key idea is to make the pooling that occurs in each convolutional
layer a stochastic process. Conventional forms of pooling such as
average and max are deterministic, the latter selecting the largest
activation in each pooling region. In our stochastic pooling, the
selected activation is drawn from a multinomial distribution
formed by the activations within the pooling region.

An alternate view of stochastic pooling is that it is equivalent to
standard max pooling but with many copies of an input image, each
having small local deformations. This is similar to explicit elastic
deformations of the input images \cite{Simard03}, which delivers
excellent MNIST performance. Other types of data augmentation, such as
flipping and cropping differ in that they are global image
transformations. Furthermore, using stochastic pooling in a
multi-layer model gives an exponential number of deformations since
the selections in higher layers are independent of those below.

%

\section{Review of Convolutional Networks}

Our stochastic pooling scheme is designed for use in a standard
convolutional neural network architecture. We first review this model,
along with conventional pooling schemes, before introducing our novel
stochastic pooling approach.

A classical convolutional network is composed of alternating layers of
convolution and pooling (i.e.~subsampling).  The aim of the first
convolutional layer is to extract patterns found within local regions
of the input images that are common throughout the dataset.  This is
done by convolving a template or filter over the input image pixels,
computing the inner product of the template at every location in the
image and outputting this as a feature map $c$, for each filter in the
layer. This output is a measure of how well the template matches each
portion of the image. A non-linear function $f()$ is then applied
element-wise to each feature map $c$: $a = f(c)$. The resulting
activations $a$ are then passed to the pooling layer. This aggregates
the information within a set of small local regions, $R$, producing a
pooled feature map $s$ (of smaller size) as output. Denoting the
aggregation function as $pool()$, for each feature map $c$ we have:
\BE s_j = pool(f(c_i)) \;\;\; \forall i \in R_j \EE where $R_j$ is
pooling region $j$ in feature map $c$ and $i$ is the index of each
element within it.

The motivation behind pooling is that the activations in the pooled map
$s$ are less sensitive to the precise locations of structures within
the image than the original feature map $c$. In a multi-layer model,
the convolutional layers, which take the pooled maps as input,
can thus extract features that are increasingly invariant to local
transformations of the input image. This is important for
classification tasks, since these transformations obfuscate the object
identity.

A range of functions can be used for $f()$, with $tanh()$ and logistic
functions being popular choices. In this is paper we use a linear rectification
function $f(c) = max(0,c)$ as the non-linearity. In
general, this has been shown \cite{Nair10} to have significant
benefits over $tanh()$ or logistic functions. However, it is especially
suited to our pooling mechanism since: (i) our formulation involves the non-negativity
of elements in the pooling regions and (ii) the clipping of negative
responses introduces zeros into the pooling regions, ensuring that the
stochastic sampling is selecting from a few specific locations (those
with strong responses), rather than all possible locations in the
region.

There are two conventional choices for $pool()$: average and max. The
former takes the arithmetic mean of the elements in each pooling
region:
\BE s_j = \frac{1}{|R_j|} \sum_{i \in R_j} a_i \:\:\:  \EE
while the max operation selects the largest element:
\BE s_j = \max_{i \in R_j} a_i
\EE

Both types of pooling have drawbacks when training deep convolutional
networks. In average pooling, all elements in a pooling region are considered,
even if many have low magnitude. When combined with linear rectification
non-linearities, this has the effect of down-weighting strong
activations since many zero elements are included in the average.
Even worse, with $tanh()$ non-linearities, strong positive and negative activations
can cancel each other out, leading to small pooled responses.

While max pooling does not suffer from these drawbacks, we find it easily overfits the training set
in practice,
making it hard to generalize well to test examples. Our proposed pooling scheme has the
advantages of max pooling but its stochastic nature helps prevent
over-fitting.

\section{Stochastic Pooling}

In stochastic pooling, we select the pooled map response by sampling
from a multinomial distribution formed from the activations of each
pooling region. More precisely, we first compute the probabilities $p$
for each region $j$ by normalizing the activations within the
region:
\BE
p_i = \frac{a_i}{\sum_{k \in R_j} a_k}
\label{eqn:multinomial}
\EE

\begin{figure}[h!]
\begin{center}
\includegraphics[width=5.5in]{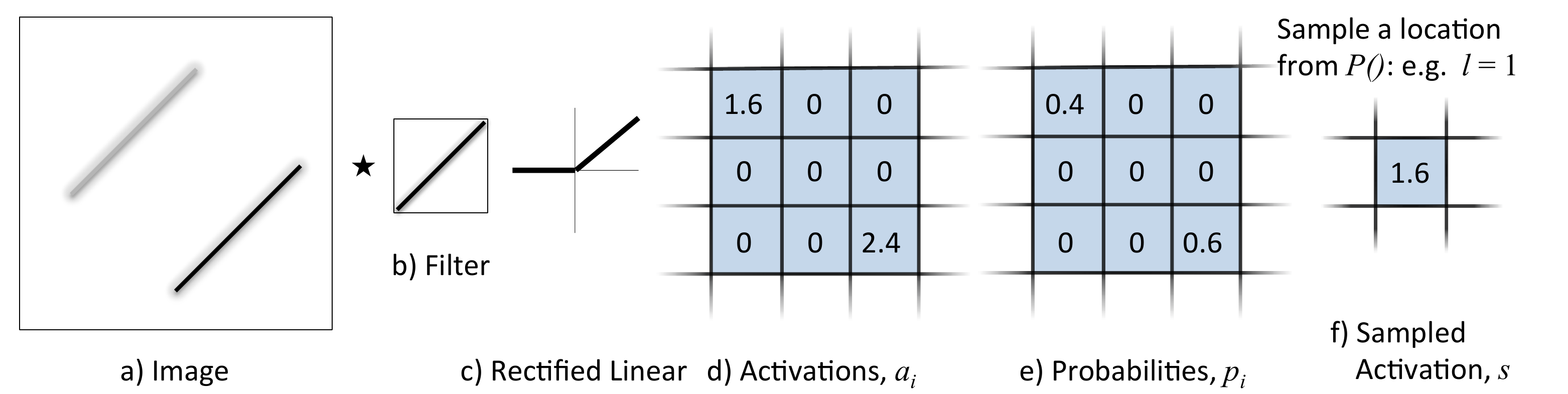}
\end{center}
\vspace*{-0.2cm}
\caption{Toy example illustrating stochastic pooling. a) Input image. b)
  Convolutional filter. c) Rectified linear function. d) Resulting activations within a given pooling region. e) Probabilities based
  on the activations. f) Sampled activation. Note that the selected
  element for the pooling region may not be the largest
  element. Stochastic pooling can thus represent multi-modal
  distributions of activations within a region.}
\label{fig:toy}
\vspace*{-0.2cm}
\end{figure}

We then sample from the multinomial distribution based on
$p$ to pick a location $l$ within the region. The pooled activation is
then simply $a_l$:
\BE
 s_j = a_l \:\:\: \text{where} \:\: l \sim P(p_1,\ldots,p_{|R_j|})
\EE
The procedure is illustrated in \fig{toy}. The samples for each
pooling region in each layer for each training example are drawn
independently to one another. When back-propagating through the network
this same selected location $l$ is used to direct the gradient back
through the pooling region, analogous to back-propagation with max pooling.

Max pooling only captures the strongest activation of the filter
template with the input for each region. However, there may be
additional activations in the same pooling region that should be taken
into account when passing information up the network and stochastic
pooling ensures that these non-maximal activations will also be utilized.


\subsection{Probabilistic Weighting at Test Time}

Using stochastic pooling at test time introduces noise into
the network's predictions which we found to degrade performance (see \secc{averaging}). Instead, we
use a probabilistic form of averaging. In this, the activations in
each region are weighted by the probability $p_i$ (see
\eqn{multinomial}) and summed:
\BE
s_j = \sum_{i \in R_j} p_i a_i
\EE
This differs from standard average pooling because each element has a potentially different
weighting and the denominator is the
sum of activations $\sum_{i \in R_j} a_i$, rather than the pooling
region size $|R_j|$. In practice, using conventional average (or sum) pooling
results in a huge performance drop (see \secc{averaging}).


Our probabilistic weighting can be viewed as a form of model averaging
in which each setting of the locations $l$ in the pooling regions
defines a new model. At training time, sampling to get new locations
produces a new model since the connection structure throughout the
network is modified. At test time, using the probabilities instead of
sampling, we effectively get an estimate of averaging over all of these possible
models without having to instantiate them.
Given a network architecture with $d$ different pooling
regions, each of size $n$, the number of possible models is $n^d$ where $d$ can be in
the $10^4$-$10^6$ range and $n$ is typically 4,9, or 16 for example
(corresponding to $2\times2$, $3\times3$ or $4\times4$ pooling
regions). This is a significantly larger number than the model averaging
that occurs in dropout \cite{Hinton12}, where $n=2$ always (since an activation is either present or
not).
In \secc{averaging} we confirm that using this
probability weighting achieves similar performance compared to using a large number of model
instantiations, while requiring only one pass through the network.

Using the probabilities for sampling at training time and for
weighting the activations at test time leads to state-of-the-art
performance on many common benchmarks, as we now demonstrate.

\section{Experiments}
\label{exp_sec}

\subsection{Overview}
We compare our method to average and max pooling on a
variety of image classification tasks. In all experiments we use
mini-batch gradient descent with momentum to optimize the cross
entropy between our network's prediction of the class and the ground
truth labels. For a given parameter $x$ at time $t$ the weight updates
added to the parameters, $\Delta x_t$ are $\Delta x_t = 0.9 \Delta
x_{t-1} - \epsilon g_t$ where $g_t$ is the gradient of the cost
function with respect to that parameter at time $t$ averaged over the
batch and $\epsilon$ is a learning rate set by hand.

\begin{figure}[t!]
\begin{center}
\includegraphics[width=5.5in]{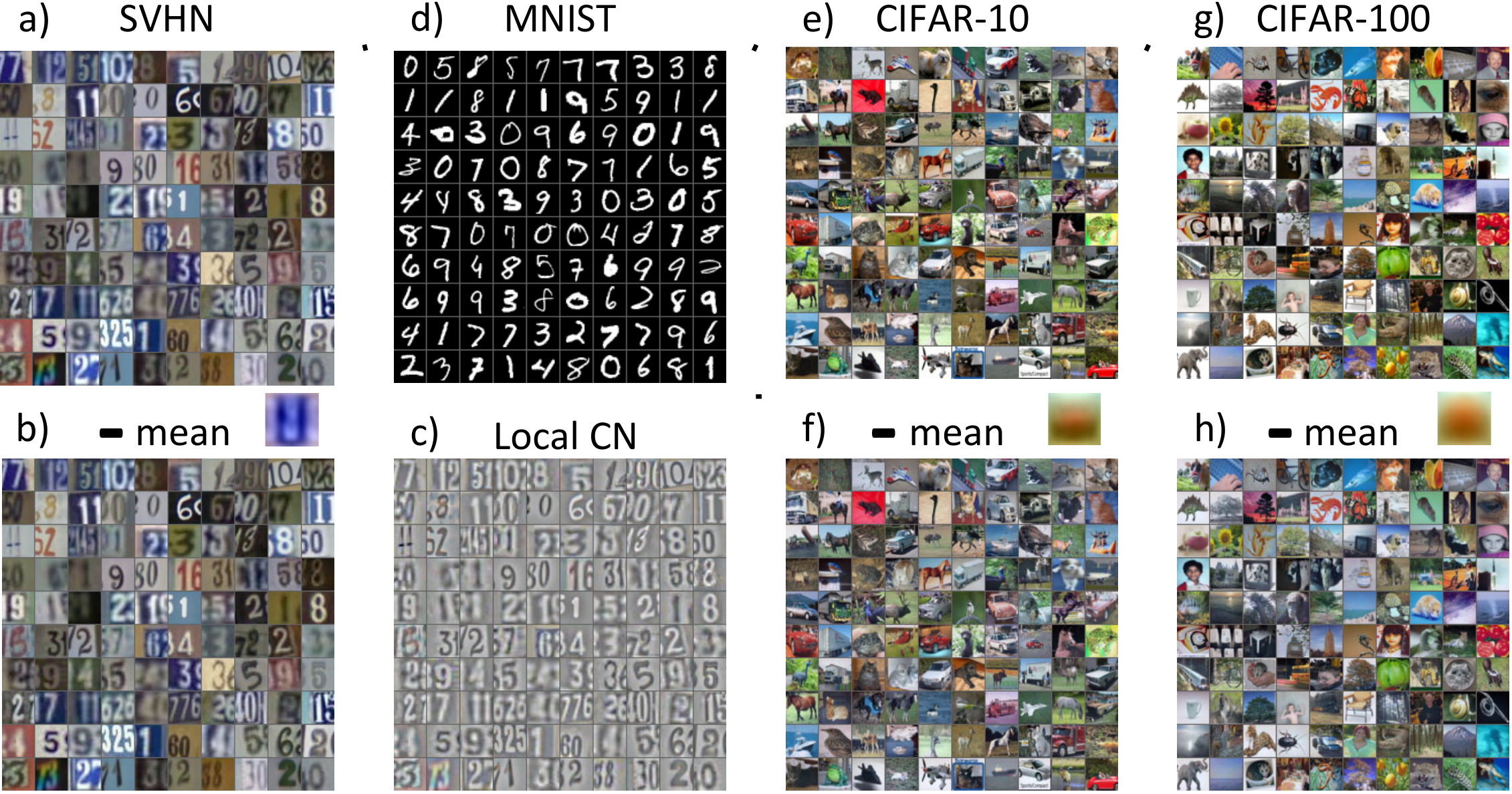}
\end{center}
\vspace*{-0.4cm}
\caption{A selection of images from each of the datasets we
  evaluated. The top row shows the raw images while the bottom row are
  the preprocessed versions of the images we used for training. The
  CIFAR datasets (f,h) show slight changes by subtracting the per
  pixel mean, whereas SVHN (b) is almost indistinguishable from the
  original images. This prompted the use of local contrast
  normalization (c) to normalize the extreme brightness variations and
  color changes for SVHN.}
\label{fig:datasets}
\vspace*{-0.4cm}
\end{figure}

All experiments were conducted using an extremely efficient C++ GPU
convolution library \cite{cudaconvnet} wrapped in MATLAB using GPUmat
\cite{gpumat}, which allowed for rapid development and
experimentation. We begin with the same network layout as in Hinton
\etal's dropout work \cite{Hinton12}, which has $3$ convolutional layers
with 5x5 filters and $64$ feature maps per layer with rectified linear units as their
outputs. We use this same model and train for 280 epochs in all experiments aside
from one additional model in \secc{SVHN} that has 128 feature maps in layer 3 and is trained for 500
epochs. Unless otherwise specified we use $3 \times 3$
pooling with stride $2$
(i.e.~neighboring pooling regions overlap by $1$ element along the
borders) for each of the $3$ pooling layers. Additionally, after each pooling layer there is a
response normalization layer (as in \cite{Hinton12}), which normalizes the
pooling outputs at each location over a subset of neighboring feature
maps. This typically helps training by suppressing extremely large
outputs allowed by the rectified linear units as well as helps
neighboring features communicate. Finally, we use a single
fully-connected layer with soft-max outputs to produce the network's
class predictions. We applied this model to four different datasets:
MNIST, CIFAR-10, CIFAR-100 and Street View House Numbers (SVHN), see
\fig{datasets} for examples images.

\subsection{CIFAR-10}
We begin our experiments with the CIFAR-10 dataset where
convolutional networks and methods such as dropout are known to work
well \cite{Hinton12,Kriz09}. This dataset is composed of 10 classes of
natural images with 50,000 training examples in total, 5,000 per
class. Each image is an RGB image of size 32x32 taken from the tiny
images dataset and labeled by hand. For this dataset we scale to [0,1] and follow Hinton \etal's \cite{Hinton12}
approach of subtracting the per-pixel mean computed over the dataset from each image as shown
in \fig{datasets}(f).

Cross-validating with a set of 5,000 CIFAR-10 training images, we
found a good value for the learning rate $\epsilon$ to be
$10^{-2}$ for convolutional layers and $1$ for the final softmax
output layer. These rates were annealed linearly throughout training
to $1/100th$ of their original values.
Additionally, we found a small weight decay of $0.001$ to be optimal
and was applied to all layers. These hyper-parameter settings found through cross-validation were
used for all other datasets in our experiments.

\begin{figure}[t!]
\vspace*{-0.3cm}
\begin{center}
\includegraphics[width=2.8in]{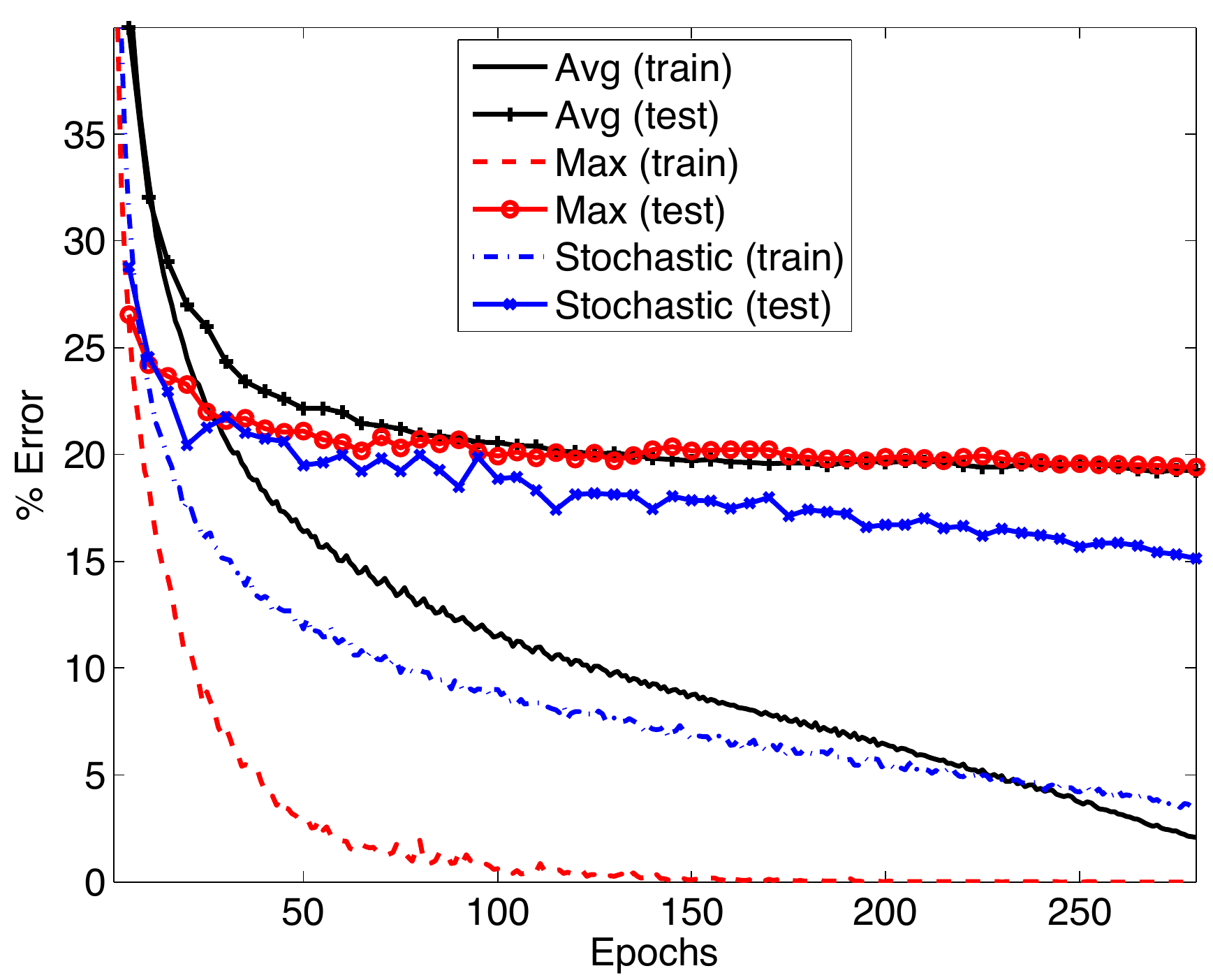}
\end{center}
\vspace*{-0.6cm}
\caption{CIFAR-10 train and test error rates throughout training for
  average, max, and stochastic pooling. Max and average pooling test
  errors plateau as those methods overfit. With stochastic pooling,
  training error remains higher while test errors continue to
  decrease.\footnotemark[1]}
\label{fig:cifar}
\vspace*{-0.3cm}
\end{figure}

Using the same network architecture described above, we trained
three models using average, max and stochastic pooling respectively and compare
their performance. \fig{cifar} shows the progression of train and test
errors over 280 training epochs. Stochastic pooling avoids
over-fitting, unlike average and max pooling, and produces less test
errors. \tab{cifar} compares the test performance of the three pooling
approaches to the current state-of-the-art result on CIFAR-10 which uses no data augmentation but
adds dropout on an additional locally connected layer
\cite{Hinton12}. Stochastic pooling surpasses this
result by 0.47\% using the same architecture but without requiring the
locally connected layer.


\footnotetext[1]{Weight decay prevented training errors from reaching
  0 with average and stochastic pooling methods and required the high number
  of epochs for training. All methods performed slightly better with weight decay.}

\begin{table}[h!]
\small
\vspace*{-2mm}
\begin{center}
\begin{tabular}{|l|c|c|}
  \hline
  & Train Error \% & Test Error \% \\
  \hline 3-layer Conv. Net  \cite{Hinton12} & -- & $16.6$ \\
  \hline 3-layer Conv. Net + 1 Locally Conn. layer with dropout \cite{Hinton12} & -- & $15.6$ \\
  \hline \hline 
  Avg Pooling & $1.92$ & $19.24$ \\
   \hline 
  Max Pooling & $0.0$ & $19.40$ \\
   \hline 
  Stochastic Pooling & $3.40$ & $\textbf{15.13}$ \\
   \hline
\end{tabular}
\vspace*{-2mm}
\caption{CIFAR-10 Classification performance for various pooling
  methods in our model compared to the state-of-the-art performance
  \cite{Hinton12} with and without dropout.}
\label{tab:cifar}
\vspace*{-2mm}
\end{center}
\end{table}

\begin{figure}[h!]
\begin{center}
\includegraphics[width=4.8in]{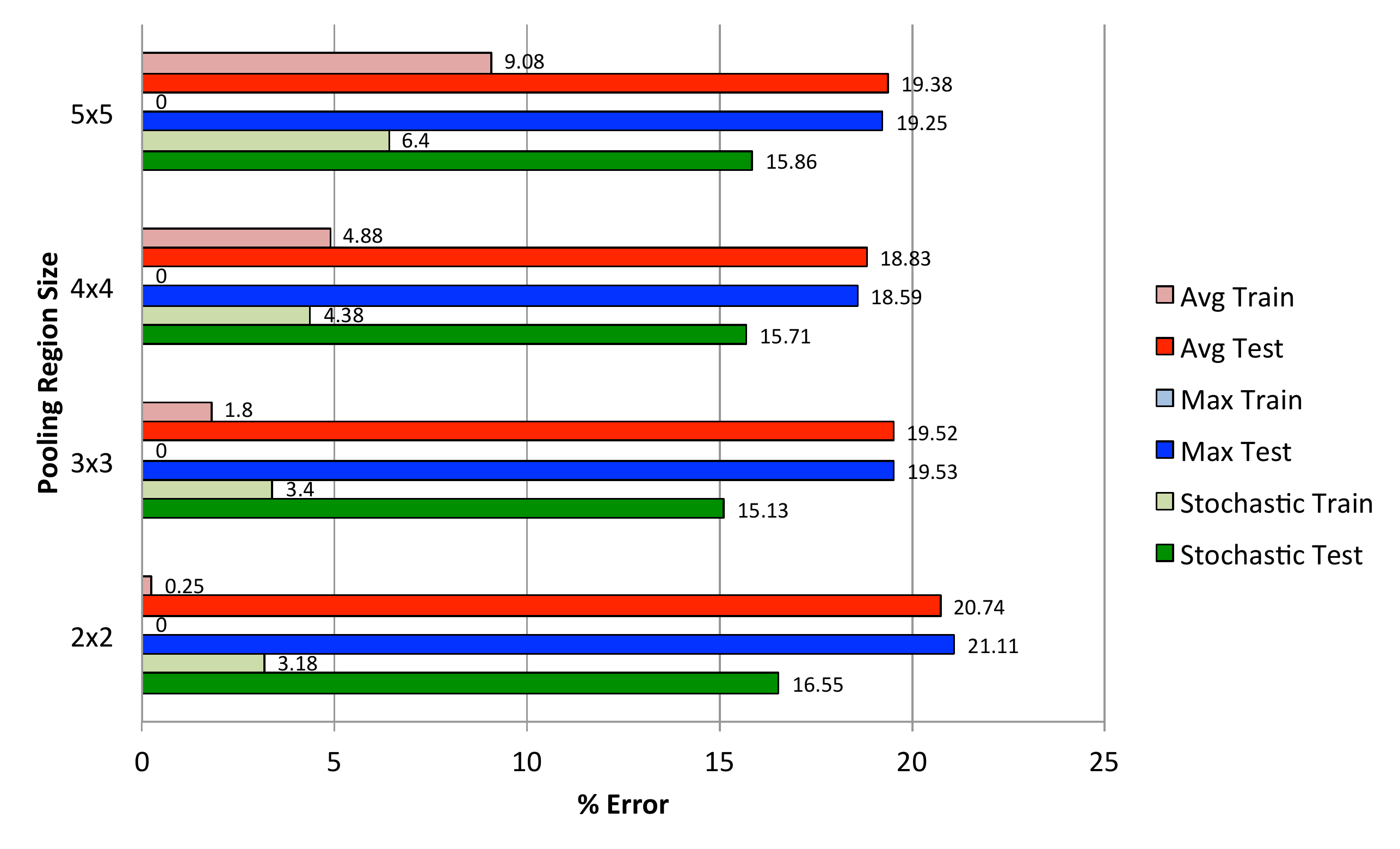}
\end{center}
\vspace*{-0.8cm}
\caption{CIFAR-10 train and test error rates for various pooling region sizes with each method.}
\label{fig:cifar_ps}
\vspace*{-0.5cm}
\end{figure}

To determine the effect of the pooling region size on the behavior of
the system with stochastic pooling, we compare the CIFAR-10 train
and test set performance for 5x5, 4x4, 3x3, and 2x2 pooling sizes
throughout the network in \fig{cifar_ps}. The optimal size appears to
be 3x3, with smaller regions over-fitting and larger regions possibly
being too noisy during training. At all sizes the stochastic pooling
outperforms both max and average pooling.

\subsection{MNIST}

The MNIST digit classification task is composed of 28x28 images of the
10 handwritten digits \cite{MNIST}. There are 60,000 training images with 10,000
test images in this benchmark. The images are scaled to [0,1] and we do not perform any other
pre-processing.


During training, the error using both stochastic pooling and max
pooling dropped quickly, but the latter completely overfit the
training data. Weight decay prevented average pooling from
over-fitting, but had an inferior performance to the other two
methods. \tab{mnist} compares the three pooling approaches to
state-of-the-art methods on MNIST, which also utilize convolutional
networks. Stochastic pooling outperforms all other methods that do not use
data augmentation methods such as jittering or elastic distortions \cite{mnistresults}.
The current state-of-the-art single model approach by Cireşan \etal
\cite{Ciresan11} uses elastic distortions to augment the
original training set. As stochastic pooling is a different type of
regularization, it could be combined with data augmentation to further
improve performance.


\begin{table}[h!]
\small
\vspace*{-2mm}
\begin{center}
\begin{tabular}{|l|c|c|}
  \hline
  & Train Error \% & Test Error \% \\
  \hline 2-layer Conv. Net + 2-layer Classifier  \cite{Jarrett2009} & -- & $0.53$ \\
  \hline 6-layer Conv. Net + 2-layer Classifier + elastic distortions \cite{Ciresan11} & -- & $0.35$ \\
  \hline  \hline   
  Avg Pooling & $0.57$ & $0.83$ \\
  \hline 
  Max Pooling & $0.04$ & $0.55$ \\
  \hline 
  Stochastic Pooling & $0.33$ & $\textbf{0.47}$ \\
  \hline
\end{tabular}
\vspace*{-2mm}
\caption{MNIST Classification performance for various pooling
  methods. Rows 1 \& 2 show the current state-of-the-art
  approaches.}
\label{tab:mnist}
\end{center}
\vspace*{-5mm}
\end{table}

\subsection{CIFAR-100}

The CIFAR-100 dataset is another subset of the tiny images dataset,
but with 100 classes \cite{Kriz09}. There are 50,000 training examples in total (500
per class) and 10,000 test examples. As with the CIFAR-10, we scale to [0,1] and subtract
the per-pixel mean from each image as shown in \fig{datasets}(h). Due
to the limited number of training examples per class, typical pooling
methods used in convolutional networks do not perform well, as shown in
\tab{cifar100}. Stochastic pooling outperforms these methods by
preventing over-fitting and surpasses what we believe to be the state-of-the-art
method by $2.66$\%.

\begin{table}[h!]
\small
\vspace*{-2mm}
\begin{center}
\begin{tabular}{|l|c|c|}
  \hline
  & Train Error \% & Test Error \% \\
  \hline Receptive Field Learning \cite{Jia11} & -- & $45.17$ \\
  \hline  \hline   
  Avg Pooling & $11.20$ & $47.77$ \\
   \hline 
  Max Pooling & $0.17$ & $50.90$ \\
   \hline 
  Stochastic Pooling & $21.22$ & $\textbf{42.51}$ \\
   \hline
\end{tabular}
\vspace*{-1mm}
\caption{CIFAR-100 Classification performance for various pooling
  methods compared to the state-of-the-art method based on receptive
  field learning.}
\label{tab:cifar100}
\end{center}
\vspace*{-6mm}
\end{table}

\subsection{Street View House Numbers} \label{sec:SVHN}

The Street View House Numbers (SVHN) dataset is composed of 604,388
images (using both the difficult training set and simpler extra set)
and 26,032 test images \cite{SVHN}. The goal of this task is to classify the digit
in the center of each cropped 32x32 color image. This is a difficult real
world problem since multiple digits may be visible within each
image. The practical application of this is to classify house numbers
throughout Google's street view database of images.

We found that subtracting the per-pixel mean from each image did not
really modify the statistics of the images (see \fig{datasets}(b)) and
left large variations of brightness and color that could make
classification more difficult. Instead, we utilized local contrast
normalization  (as in \cite{Sermanet11}) on each of the three RGB channels to pre-process the
images \fig{datasets}(c). This normalized the brightness and color
variations and helped training proceed quickly on this relatively
large dataset.

Despite having significant amounts of training data, a large
convolutional network can still overfit. For this dataset, we train an additional model for 500
epochs with 64, 64 and 128 feature maps in layers 1, 2 and 3 respectively. Our stochastic pooling
helps to prevent overfitting even in this large model (denoted 64-64-128 in \tab{svhn}), despite
training for a long time. The existing
state-of-the-art on this dataset is the multi-stage
convolutional network of Sermanet \etal \cite{Sermanet11}, but
stochastic pooling beats this by $2.10$\% (relative gain of $43\%$).

\begin{table}[h!]
\small
\vspace*{-2mm}
\begin{center}
\begin{tabular}{|l|c|c|}
  \hline
  & Train Error \% & Test Error \% \\
  \hline Multi-Stage Conv. Net + 2-layer Classifier  \cite{Sermanet11} & -- & $5.03$ \\
  \hline Multi-Stage Conv. Net + 2-layer Classifer + padding \cite{Sermanet11} & -- & $4.90$ \\
  \hline \hline   
  64-64-64 Avg Pooling  & $1.83$ & $3.98$ \\
  \hline 
  64-64-64 Max Pooling  & $0.38$ & $3.65$ \\
  \hline 
  64-64-64 Stochastic Pooling  & $1.72$ & $3.13$ \\ 
  \hline \hline   
  64-64-128 Avg Pooling  & $1.65$ & $3.72$ \\
  \hline 
  64-64-128 Max Pooling  & $0.13$ & $3.81$ \\
  \hline 
  64-64-128 Stochastic Pooling  & $1.41$ & $\textbf{2.80}$ \\ 
  \hline
\end{tabular}
\vspace*{-2mm}
\caption{SVHN Classification performance for various pooling methods in
  our model with 64 or 128 layer 3 feature maps compared to state-of-the-art results with and
  without data augmentation.}
\label{tab:svhn}
\vspace*{-6mm}
\end{center}
\end{table}

\subsection{Reduced Training Set Size}
\begin{figure}[h!]
\vspace*{-0.4cm}
\begin{center}
\mbox{
\includegraphics[width=2.5in]{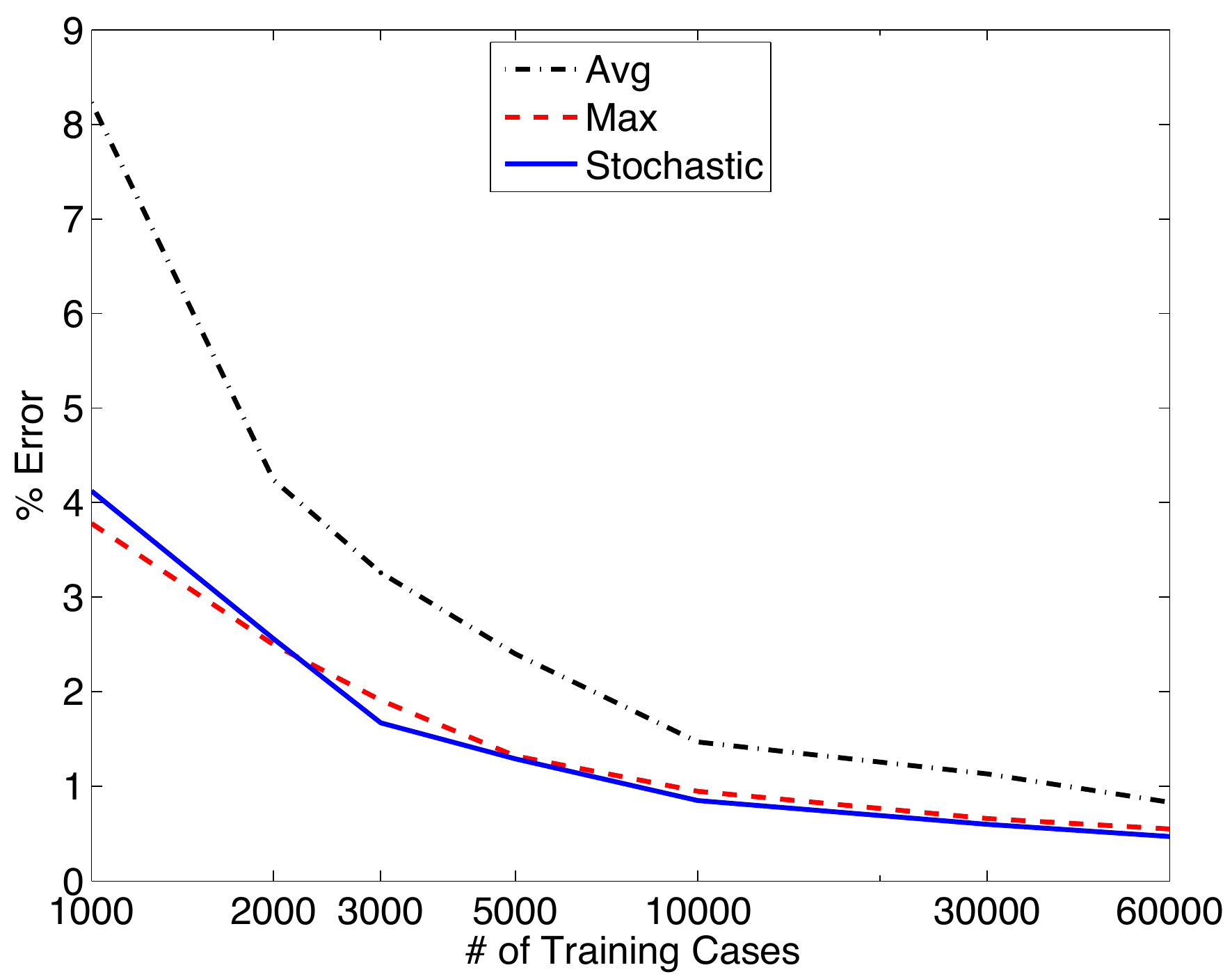}
\includegraphics[width=2.5in]{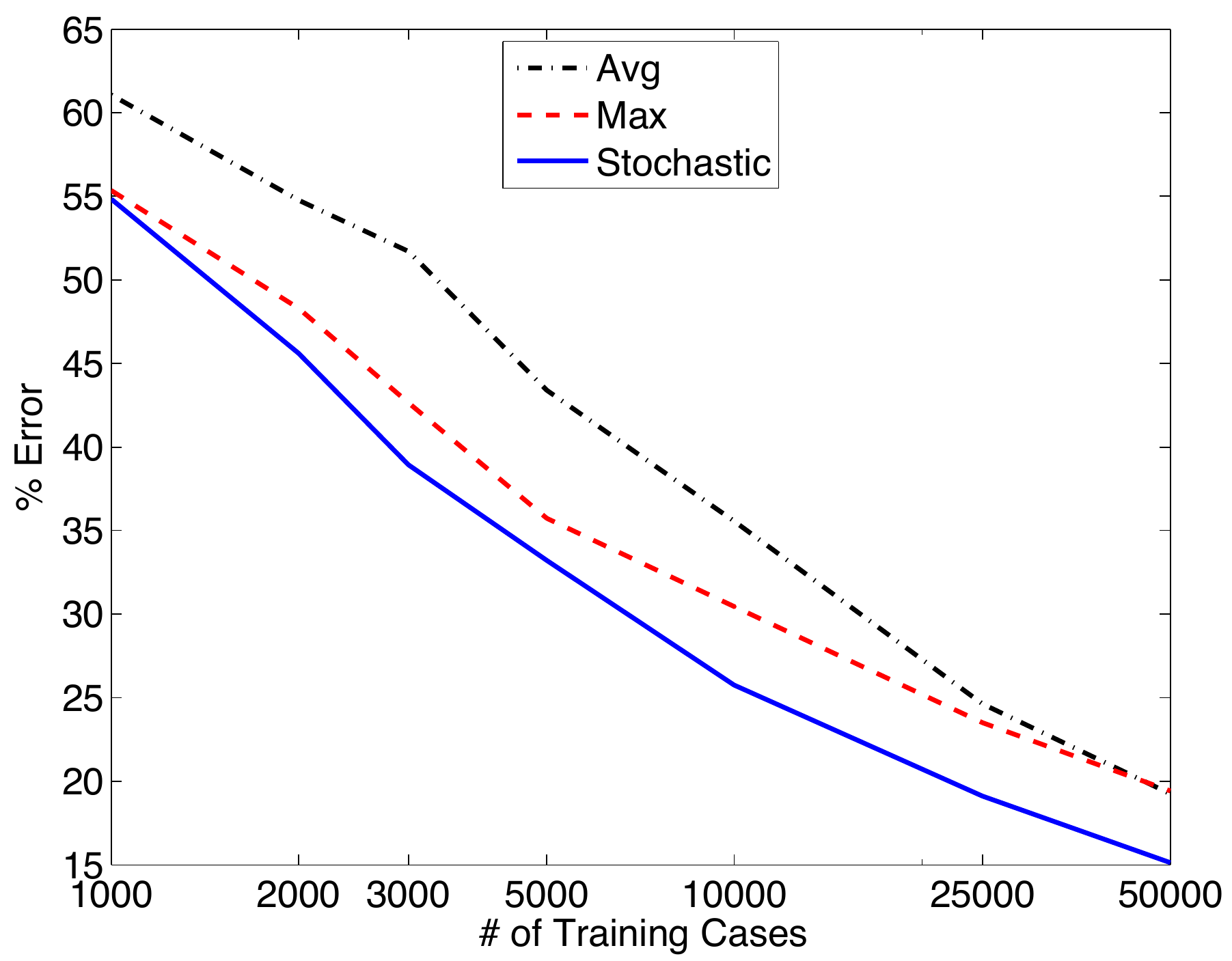}}
\end{center}
\vspace*{-0.6cm}
\caption{Test error when training with reduced dataset sizes on MNIST (left) and
  CIFAR-10 (right). Stochastic pooling generally overfits the least. }
\label{fig:reduced}
\vspace*{-0.4cm}
\end{figure}

\vspace{-1mm}
To further illustrate the ability of stochastic pooling to prevent
over-fitting, we reduced the training set size on MINST and CIFAR-10
datasets. \fig{reduced} shows test performance when training on a
random selection of only 1000, 2000, 3000, 5000, 10000, half, or the
full training set. In most cases, stochastic pooling overfits less than the other pooling
approaches.

\subsection{Importance of Model Averaging} \label{sec:averaging}
\vspace{-1mm}
To analyze the importance of stochastic sampling at training time and
probability weighting at test time, we use different methods of
pooling when training and testing on CIFAR-10 (see
\tab{combos}). Choosing the locations stochastically at test time
degrades performance slightly as could be expected, however it still
outperforms models where max or average pooling are used at test
time. To confirm that probability weighting is a valid approximation to averaging many models, we draw $N$ samples
of the pooling locations throughout the network and average the output probabilities from those $N$
models (denoted Stochastic-$N$ in \tab{combos}). As $N$ increases, the results approach the
probability weighting method, but have the obvious downside of an $N$-fold increase in computations.

Using a model trained with max or average
pooling and using stochastic pooling at test time performs
poorly. This suggests that training with stochastic pooling, which
incorporates non-maximal elements and sampling noise, makes the model
more robust at test time. Furthermore, if these non-maximal elements
are not utilized correctly or the scale produced by the pooling
function is not correct, such as if average pooling is used at test
time, a drastic performance hit is seen.

When using probability weighting during training, the network easily
over-fits and performs sub-optimally at test time using any of the
pooling methods. However, the benefits of probability
weighting at test time are seen when the model has specifically
been trained to utilize it through either probability weighting or
stochastic pooling at training time.

\begin{table}[h!]
\small
\vspace*{-3mm}
\begin{center}
\begin{tabular}{|l|c|c|c|}
  \hline
  Train Method & Test Method & Train Error \% & Test Error \% \\
  \hline\hline
  Stochastic Pooling & Probability Weighting & $3.20$ & $\textbf{15.20}$ \\
  Stochastic Pooling & Stochastic Pooling & $3.20$ & $17.49$ \\
  Stochastic Pooling & Stochastic-10 Pooling & $3.20$ & $15.51$ \\
  Stochastic Pooling & Stochastic-100 Pooling & $3.20$ & $\textbf{15.12}$ \\
  Stochastic Pooling & Max Pooling & $3.20$ & $17.66$ \\ 
  Stochastic Pooling & Avg Pooling & $3.20$ & $53.50$ \\ 
  \hline\hline 
  Probability Weighting & Probability Weighting & $0.0$ & $19.40$ \\
  Probability Weighting & Stochastic Pooling & $0.0$ & $24.00$ \\
  Probability Weighting & Max Pooling & $0.0$ & $22.45$ \\
  Probability Weighting & Avg Pooling & $0.0$ & $58.97$ \\
  \hline\hline 
  Max Pooling & Max Pooling & $0.0$ & $19.40$ \\
  Max Pooling & Stochastic Pooling & $0.0$ & $32.75$ \\
  Max Pooling & Probability Weighting & $0.0$ & $30.00$ \\
  \hline\hline 
  Avg Pooling & Avg Pooling & $1.92$ & $19.24$ \\
  Avg Pooling & Stochastic Pooling & $1.92$ & $44.25$ \\
  Avg Pooling & Probability Weighting & $1.92$ & $40.09$ \\
  \hline
\end{tabular}
\vspace*{-2mm}
\caption{CIFAR-10 Classification performance for various train and
  test combinations of pooling methods. The best performance is
  obtained by using stochastic pooling when training (to prevent
  over-fitting), while using the probability weighting at test time.}
\label{tab:combos}
\end{center}
\vspace*{-9mm}
\end{table}

\subsection{Visualizations}
\begin{figure}[h!]
\vspace*{-0.3cm}
\begin{center}
\includegraphics[width=5.5in]{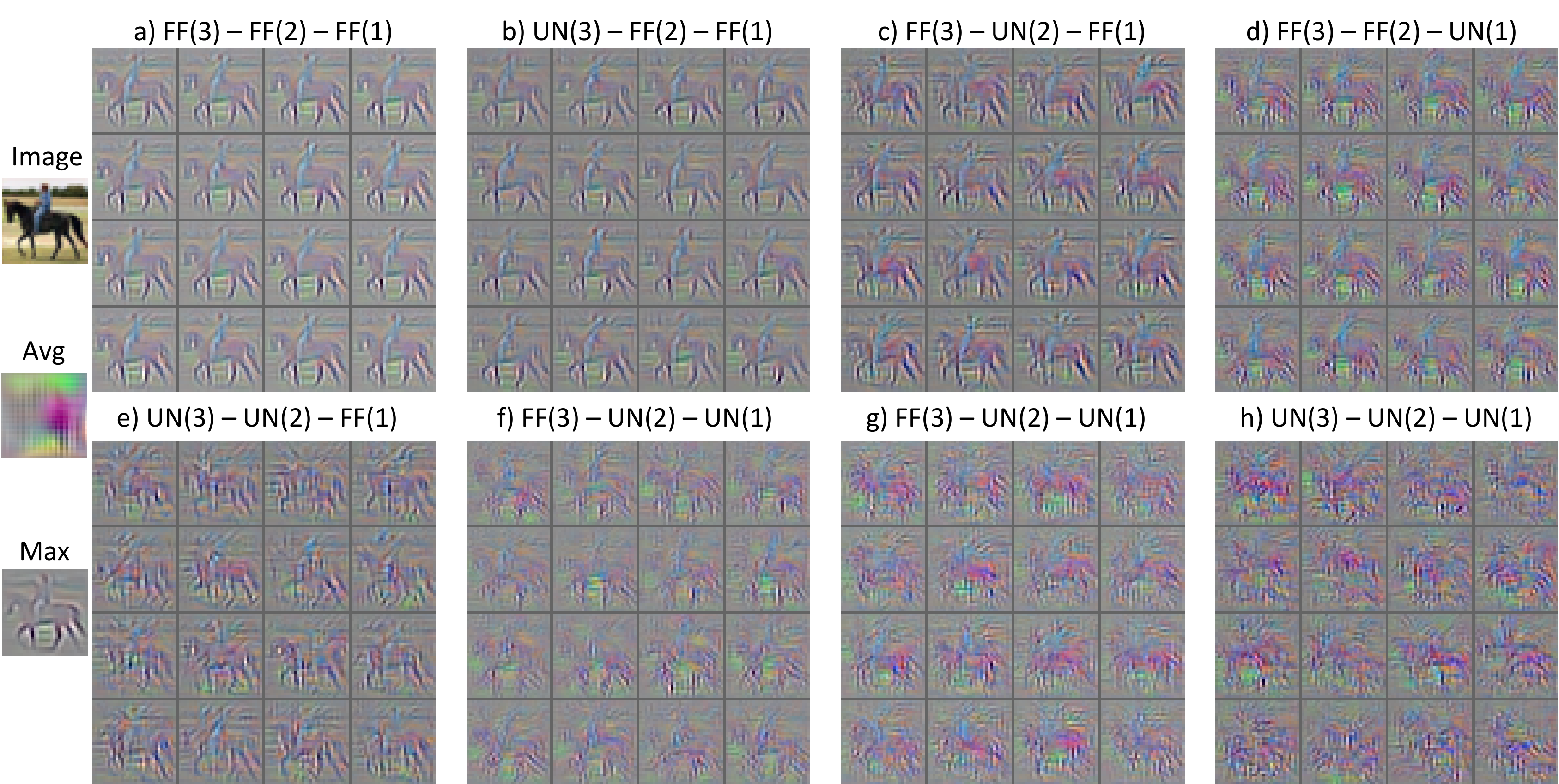}
\end{center}
\vspace*{-0.5cm}
\caption{Top down visualizations from the third layer feature map
  activations for the horse image (far left). Max and average pooling
  visualizations are also shown on the left. (a)--(h): Each image in a
  4x4 block is one instantiation of the pooling locations using
  stochastic pooling. For sampling the locations, each layer
  (indicated in parenthesis) can either use: (i) the multinomial
  distribution over a pooling region derived from the feed-forward
  (FF) activations as in \eqn{multinomial}, or (ii) a uniform (UN)
  distribution. We can see that the
  feed-forward probabilities encode much of the structure in the
  image, as almost all of it is lost when uniform sampling is used, especially in
  the lower layers. }
\label{fig:viscifar}
\vspace*{-3mm}
\end{figure}

\vspace{-2mm}
Some insight into the mechanism of stochastic pooling can be gained by using a deconvolutional
network of Zeiler \etal \cite{Zeiler11} to provide a novel visualization of our trained
convolutional network. The deconvolutional network has the same components (pooling, filtering) as a
convolutional network but are inverted to act as a top-down decoder that maps the top-layer feature
maps back to
the input pixels. The unpooling operation uses the stochastically chosen locations selected during
the forward pass. The deconvolution network filters (now applied to the feature maps, rather than
the input) are the transpose of the feed-forward filters, as in an auto-encoder with tied
encoder/decoder weights. We repeat this top-down process until the input pixel level is reached, producing
the visualizations in \fig{viscifar}. With max pooling, many of the input image edges are present,
but average pooling produces a reconstruction with no discernible structure.  \fig{viscifar}(a)
shows 16 examples of pixel-space reconstructions for different location samples throughout the
network. The reconstructions are similar to the max pooling case, but as the pooling locations
change they result in small local deformations of the visualized image.

Despite the stochastic nature of the model, the multinomial distributions
effectively capture the regularities of the data. To demonstrate this, we compare the outputs
produced by a deconvolutional network when sampling using the feedforward (FF) proabilities versus
sampling from uniform (UN) distributions.
In contrast to \fig{viscifar}(a) which uses only feedforward proabilities, \fig{viscifar}(b-h)
replace one or more of the pooling layers' distributions with uniform distributions.
The feed forward
probabilities encode significant structural information, especially in
the lower layers of the model.  Additional visualizations and videos of
the sampling process are provided as supplementary material at
\url{www.matthewzeiler.com/pubs/iclr2013/}.


\vspace{-4mm}
\section{Discussion}
\vspace{-3mm}
We propose a simple and effective stochastic pooling
strategy that can be combined with any other forms of regularization
such as weight decay, dropout, data augmentation, etc.~to prevent
over-fitting when training deep convolutional networks. The method is
also intuitive, selecting from information the network is already
providing, as opposed to methods such as dropout which throw
information away. We show state-of-the-art performance on numerous
datasets, when comparing to other approaches that do not employ data
augmentation. Furthermore, our method has negligible computational
overhead and no hyper-parameters to tune, thus can be swapped into to
any existing convolutional network architecture.

\baselineskip=1pt
\bibliographystyle{plain}
\bibliography{iclr2013}
\end{document}